\documentclass{article} 
\usepackage{nips10submit_e,times}
\usepackage{graphicx}
\usepackage{amsmath,amsfonts}
\usepackage{hyperref}

\title{Brain covariance selection: better individual functional
connectivity models using population prior}

\author{
Ga\"el Varoquaux$^{\star}$\\
Parietal, INRIA\\
NeuroSpin, CEA, France\\
\texttt{gael.varoquaux@normalesup.org} \\
\And
Alexandre Gramfort\\
Parietal, INRIA\\
NeuroSpin, CEA, France\\
\texttt{alexandre.gramfort@inria.fr}\\
\And
Jean-Baptiste Poline\\
LNAO, I2BM, DSV\\
NeuroSpin, CEA, France\\
\texttt{jbpoline@cea.fr} \\
\And
Bertrand Thirion\\
Parietal, INRIA\\
NeuroSpin, CEA, France\\
\texttt{bertrand.thirion@inria.fr} \\
}

\providecommand{\OO}[1]{\mathcal{O}\bigl(#1\bigr)}
\providecommand{\B}[1]{\mathbf{#1}}

\nipsfinalcopy 

\begin{document}

\maketitle

\begin{abstract} 
Spontaneous brain activity, as observed in functional neuroimaging,
has been shown to display reproducible structure that expresses
brain architecture and carries markers of brain pathologies.
An important view of modern neuroscience is that such
large-scale structure of coherent activity
reflects modularity properties of brain connectivity graphs.
However, to date, there has been no demonstration that 
the limited and noisy data available in spontaneous activity 
observations could be used to learn full-brain
probabilistic models that generalize to new data. 
Learning such models entails two main challenges:
\textit{i)} modeling full brain
connectivity is a difficult estimation problem that faces the curse of
dimensionality and \textit{ii)} variability between subjects, coupled
with the variability of functional signals between experimental runs,
makes the use of multiple datasets challenging.
We describe subject-level brain functional connectivity structure as a
multivariate Gaussian process and introduce a new strategy to estimate
it from group data, by imposing a common structure on the graphical
model in the population.
We show that individual models learned from functional Magnetic
Resonance Imaging (fMRI) data using this population prior generalize
better to unseen data than models based on alternative regularization
schemes.
To our knowledge, this is the first report of a cross-validated model
of spontaneous brain activity.
Finally, we use the estimated graphical model to
explore the large-scale characteristics of functional architecture and
show for the first time that known cognitive networks appear as the
integrated communities of functional connectivity graph.
\end{abstract}

{\def\thefootnote{$\star$}
\footnotetext{Funding from INRIA-INSERM collaboration and 
grant /ANR/-08-BLAN-0250-02 VIMAGINE }
}

\section{Introduction}

The study of brain functional connectivity, as revealed through
distant correlations in the signals measured by functional Magnetic
Resonance Imaging (fMRI), represents an easily accessible, albeit
indirect marker of brain functional architecture; in the
recent years, it has given rise to fundamental insights on brain
organization by representing it as a modular graph with large
functionally-specialized networks
\cite{Fox2007,bullmore2009,smith2009}.

Among other features, the concept of functionally-specialized
cognitive network has emerged as one of the leading views in current
neuroscientific studies: regions that activate simultaneously,
spontaneously or as an evoked response, form an integrated network
that supports a specific cognitive function \cite{Fox2007,smith2009}.
In parallel, graph-based statistical analysis have shown that the
graphical models that naturally represent the correlation structure of
brain signals exhibit small-world properties: any two regions of the
brain can be connected through few intermediate steps, despite the
fact that most nodes maintain only a few direct connections
\cite{achard2006, bullmore2009}.
These experimental results are consistent with the view that 
the local neuronal systems in the brain group together to form large-scale 
distributed networks \cite{sporns2004}.
However, the link between large-scale networks corresponding to a
known cognitive function and segregation into functional connectivity
subgraphs has never been established.

At the individual level, the different brain functional networks are
attractive as their coherence, as manifested in their correlation
structure, appears impacted by brain pathologies, such as
schizophrenia \cite{cecchi2009}, neurodegenerative diseases
--e.g. Alzheimer's disease--\cite{seeley2009,huang2009}, or in the
study of brain lesions \cite{varoquaux2010b}. 
From the clinical standpoint, there is a strong interest in
spontaneous-activity data to study and diagnose brain pathologies because they
can be recorded even on severely impaired subjects \cite{greicius2008b}.

FMRI is the tool of choice to study large-scale functional connectivity, 
as it relies
on wide expertise gained through decades of brain mapping, and MRI
scanners are widely available in brain research institutes and hospitals.
However neural activity is observed in fMRI indirectly, at a limited
spatiotemporal resolution ($(3mm)^3\times3s$ typically), and is
confounded by measurement and physiological noise (cardiac and
respiratory cycles, motion).
%
%
For clinical applications as well as inference of brain fundamental
architecture, the quantitative characterization of spontaneous
activity has to rely on a probabilistic model of the signal.
The question of the robustness of covariance estimation procedures to
observation noise as well as inter-individual variability is thus 
fundamental, and has not been
addressed so far.

The focus of this work is the estimation of a large-scale Gaussian
model to give a probabilistic description of brain functional signals. 
The difficulties are two-fold: on the one hand, there is a shortage of 
data to learn a good 
covariance model from an individual subject, and on the other hand,
subject-to-subject variability poses a serious challenge to the use of
multi-subject data: this concerns the creation of population-level
connectivity templates, the estimation of the normal variability
around this template, and the assessment of non-normal variability.
In this paper, we provide evidence that
optimal regularization schemes can be used in the covariance
estimation problem, making it possible to pull data from several 
subjects. We show
that the resulting covariance model yields easily interpretable
structures, and in particular we provide the first experimental
evidence that the functionally integrated communities of brain
connectivity graphs correspond to known cognitive networks.
To our knowledge, this is the first experiment that
assesses quantitatively the goodness of fit of a full-brain functional
connectivity model to new data.
For this purpose, we introduce an unbiased cross-validation scheme
that tests the generalization power of the inferred model.

Although the proposed framework shares with so-called effective
connectivity models (SEM \cite{McIntosh1994}, DCM
\cite{daunizeau2009}) the formulation in terms of graphical model, it
is fundamentally different in that these approaches are designed to
test the coefficients of (small) graphical models in a
hypothesis-driven framework, while our approach addresses the
construction of large-scale model of brain connectivity that might be
valid at the population level, and is completely data-driven.
\cite{Honorio2010} have applied with success a similar framework to
modeling task-driven brain activity.

The layout of the paper is the following.
We first formulate the problem of estimating a high-dimensional
Gaussian graphical model from multi-subject data. Second, we detail
how we extract activity time-series for various brain regions from
fMRI data. 
Then, we compare the generalization performance of different
estimators based on various regularization procedures. Finally, we
study the graph communities of the learnt connectivity model as well as
the integration and segregation processes between these communities.
The present work opens the way to a systematic use of Gaussian
graphical Models for the analysis of functional connectivity data.

%
%



\section{Theoretical background: estimating Gaussian graphical models}

From a statistical estimation standpoint, 
the challenge to address is to estimate a covariance or a correlation
matrix giving a good description of the brain activation data. We choose
to use the framework of Gaussian models as these are the processes with
the minimum information --i.e. the maximum entropy-- given a covariance
matrix.


\paragraph{Covariance selection procedures}
Let us consider a dataset $\B{X} \in \mathbb{R}^{n \times p}$ with $p$
variables and $n$ samples, modeled as centered multivariate Gaussian
process. 
Estimating its covariance matrix is a difficult statistical problem
for two reasons. First, to specify a valid multivariate Gaussian
model, this covariance has to be positive definite. Second,
if $n < \frac{1}{2}p(p+1)$, as this is the case in our problem, the
number of unknown parameters is greater than the number of samples.
As a result, the eigenstructure of the sample covariance matrix
carries a large estimation error. To overcome these challenges,
Dempster \cite{Dempster1972} proposed covariance selection: learning
or setting conditional independence between variables improves the
conditioning of the problem. 
In multivariate Gaussian models, conditional independence between
variables is given by the zeros in the precision (inverse
covariance) matrix $\B{K}$. Covariance selection can thus be achieved
by imposing a sparse support for the estimated precision matrix, i.e.,
a small number of non-zero coefficients. In terms of graphical models,
this procedure amounts to limiting the number of edges.

Selecting the non-zero coefficients to optimize the likelihood of the
model given the data is a difficult combinatorial optimization problem.
It is NP hard in the number of edges.
In order to tackle this problem with more than tens of variables, it can be 
relaxed into a convex problem using a penalization
based on the $\ell_1$ norm of the precision matrix, that is known to
promote sparsity on the estimates~\cite{banerjee2006}. The optimization
problem is given by:
\begin{equation}
    \hat{\B{K}}_{\ell_1} = \text{argmin}_{\B{K} \succ 0}
                \text{tr}\, (\B{K} \, \hat{\B{\Sigma}}_\text{sample})
                - \log \det \B{K}
                + \lambda \|\B{K}\|_1,
    \label{eq:l1}
\end{equation}
where $\hat{\B{\Sigma}}_\text{sample}=\frac{1}{n}\B{X}^{T}\B{X}$ is the
sample covariance matrix, and $\|\cdot\|_1$ is the element-wise
$\ell_1$ norm of the off-diagonal coefficients in the matrix.
Optimal solutions to this
problem can be computed very efficiently in $\OO{p^3}$ time
\cite{banerjee2006, duchi2008, Friedman2008}. Note that this formulation
of the problem amounts to the computation of a maximum a posteriori
(MAP) with an i.i.d. Laplace prior on the off-diagonal coefficients of
the precision matrix.

\paragraph{Imposing a common sparsity structure}
In the application targeted by this contribution, the problem is to
estimate the precision matrices in a group of subjects among which one
can assume that all the individual precision matrices share the same
structure of conditional independence, i.e., the zeros in the different
precision matrices should be at the same positions. This amounts to a
joint prior that can also lead to the computation of a MAP.
To achieve the estimation with the latter constraint, a natural solution
consists in estimating all matrices jointly. Following the idea of
joint feature selection using the group-Lasso
for regression problems~\cite{yuan2006}, the
solution we propose consists in penalizing precisions using a mixed norm
$\ell_{21}$. Let us denote $\B{K}^{(s)}$ the
precision for subject $s$ in a population of $S$ subjects.
The penalty can be written as
$\sum_{i\neq j} \sqrt{\sum_{s=1}^S (\B{K}_{ij}^{(s)})^2} =
\sum_{i\neq j} \|\B{K}_{ij}^{(\cdot)}\|_2$.
This leads to the minimization problem:
\begin{equation}
    \left( \hat{\B{K}}_{\ell_{21}}^{(s)} \right)_{s=1..S} =
    \text{argmin}_{\B{K}^{(s)} \succ 0} \left( \sum_{s=1}^S \left(
    \text{tr}(\B{K}^{(s)} \, \hat{\B{\Sigma}}^{(s)}_\text{sample}) -
    \log \det \B{K}^{(s)} \right) + \lambda \sum_{i\neq j}
    \|\B{K}_{ij}^{(\cdot)}\|_2 \right )
\label{eq:l21}
\end{equation}
One can notice then that in the special case where $S=1$, \eqref{eq:l21} is
equivalent to \eqref{eq:l1}. By using such a penalization, a group of
coefficients $\{\hat{\B{K}}_{ij}^{(s)},\,s=1,\dots,S\}$ are either jointly set to zero
or are jointly non-zero~\cite{yuan2006}, thus one enforces the precisions
matrices to have a common sparse support for all subjects.

To our knowledge, two other recent contributions address the problem of
jointly estimating multiple graphical models \cite{guo2009,chiquet2010}.
While the approach of \cite{guo2009} is different from~\eqref{eq:l21}
and does not correspond to a group-Lasso formulation, \cite{chiquet2010}
mentions the problem~\eqref{eq:l21}. Compared to this prior work, the
optimization strategy we introduce largely differs, but also the
application and the validation settings. Indeed, we are not interested in
detecting the presence or the absence of edges on a common graph, but in
improving the estimation of a probabilistic model of the individual data.
Also, the procedure to set regularization parameter $\lambda$ is done
by evaluating the likelihood of unseen data in a principled nested
cross-validation setting.

In order to minimize~\eqref{eq:l21}, we modified the SPICE
algorithm~\cite{Rothman2008} that consists in upper bounding the
non-differentiable absolute values appearing in the $\ell_1$ norm with
a quadratic differentiable function.  When using a group-Lasso
penalty, similarly the non-differentiable $\ell_2$ norms appearing in
the $\ell_{21}$ penalty can be upper bounded. The computational
complexity of an iteration that updates all coefficients once is now
in $\OO{S \,p^3}$: it scales linearly with the number of models to 
estimate.
Following the derivation
from~\cite{duchi2008}, the iterative optimization procedure is stopped
using a condition on the optimality of the solution using a control on
the duality gap. Global optimality of the estimated solution is made
possible by the convexity of the problem~\eqref{eq:l21}.


%

Alternatively, a penalization
based on a squared $\ell_2$ norm has been investigated. It consists in
regularizing the estimate of the precision matrix by adding a diagonal
matrix to the sample covariance before computing its inverse. It
amounts to an $\ell_2$ shrinkage by penalizing uniformly off-diagonal
terms:
\begin{equation}
    \hat{\B{K}}_{\ell_2} = ( \hat{\B{\Sigma}}_\text{sample} + \lambda \,\B{I})^{-1}
\label{eqn:k_diagonal}
\end{equation}
Although the penalization parameter $\lambda$ for this shrinkage can be 
chosen by cross-validation, Ledoit and Wolf \cite{ledoit2004} have 
introduced a closed formula that leads to a good choice in practice.
Unlike $\ell_1$ penalization, $\ell_2$ downplays uniformly connections
between variables, and is thus of less interest for the study of brain
structure. It is presented mainly for comparison purposes.

\section{Probing brain functional covariance with fMRI}

\paragraph{Inter-individual variability of resting-state fMRI}
We are interested in modeling spontaneous brain activity, also called
{\sl resting state} data, recorded with fMRI. Although such data
require complex strategies to provide quantitative information on
brain function, they are known to reveal intrinsic features of brain
functional anatomy, such as cognitive networks
\cite{Fox2007,beckmann2004,smith2009} or connectivity
topology \cite{achard2006,bullmore2009}. 

A well-known challenge with brain imaging data is that no two brains
are alike. Anatomical correspondence between subjects is usually
achieved by estimating and applying a deformation field that maps the
different anatomies to a common template.
In addition to anatomical variability, within a population of
subjects, cognitive networks may recruit slightly different regions.
Our estimation strategy is based on the hypothesis that
although the strength of correlation between connected brain region
may vary across subjects, many of the conditional independence
relationship will be preserved, as they reflect the structural wiring.

\paragraph{The data at hand: multi-subject brain activation time series}
20 healthy subjects were scanned twice in a resting task, eyes closed,
resulting in a set of 244 brain volumes per session acquired with a
repetition time of 2.4 s. As in \cite{huang2009}, after standard
neuroimaging pre-processing, we extract brain fMRI time series and
average them based on an atlas that subdivides the gray matter
tissues into standard regions.

We have found that the choice of the atlas used to extract time-series is
crucial. Depending on whether the atlas oversegments brain lobes 
into regions smaller than subject-to-subject anatomical variability
or captures this variability, cross-validation scores vary
significantly. 
Unlike previous studies \cite{achard2006,huang2009}, we choose to rely
on an inter-subject probabilistic atlas of anatomical structures. 
For cortical structures, we use the prior probability of cortical
folds in template space\footnote{The corresponding atlas can be
downloaded on \url{http://lnao.lixium.fr/spip.php?article=229}} used in Bayesian sulci labeling and
normalization of the cortical surface \cite{perrot2009}. This atlas
covers 122 landmarks spread throughout the whole cortex and matches
naturally their anatomical variability in terms of position, shape,
and spread.
It has been shown to be a good support to define regions of interest
for fMRI studies \cite{keller2009}. For sub-cortical structures, such
as gray nuclei, we use the Harvard-Oxford sub-cortical probabilistic
atlas, as shipped by the FSL software package. The union of both
atlases forms an inter-subject probabilistic atlas for 137
anatomically-defined regions.

As we are interested in modeling only gray-matter correlations, we
regress out confound effects obtained by extracting signals in different
white matter and cortico-spinal fluid (CSF) regions, as well as the
rigid-body motion time courses estimated during data pre-processing. 
We use the SPM software to derive voxel-level tissue probability of
gray matter, white matter, and CSF from the anatomical images of each
subject. Tissue-specific time series for either confound signals or
grey-matter signals are obtained by multiplying the subject-specific
tissue probability maps with the probabilistic atlas.

Finally, as the fMRI signals contributing to functional connectivity
have been found to lie in frequencies below 0.1 Hz \cite{cordes2000},
we apply temporal low-pass filtering to the extracted time series. We
set the cut-off frequency of the filter using cross-validation with
the Ledoit-Wolf $\ell_2$-shrinkage estimator. We
find an optimal choice of 0.3 Hz. Also, we remove residual linear
trends due to instrument bias or residual movement signal and normalize
the variance of the resulting time series. The covariance matrices that
we study thus correspond to correlations.


\section{Learning a better model for a subject's spontaneous activity}

\paragraph{Model-selection settings}
Given a subject's resting-state fMRI dataset, our goal is to estimate
the best multivariate normal model describing this subject's functional
connectivity. 
For this, we learn the model using the data from one session, and measure the
likelihood of the second session's data from the same subject.
%
%
We use this two-fold cross-validation procedure to tune the
regularization parameters. In addition, we can use the data of the
remaining subjects as a reference population during the training
procedure to inform the model for the singled-out subject.

\paragraph{Generalization performance for different estimation strategies}
We compare different estimation strategies. First, we learn the model
using only the subject's data. We compare the sample correlation matrix,
as well as the Ledoit-Wolf, $\ell_2$ and $\ell_1$-penalized estimators. 
Second, we use the combined data of the subject's training session as
well as the population, using the same estimators: we concatenate the
data of the population and of the train session to estimate the
covariance.
Finally, we use the
$\ell_{21}$-penalized estimator in Eq.(\ref{eq:l21}), to learn
different precisions for each subject, with a common sparse structure.
As this estimation strategy yields a different correlation matrix for
each subject, we use the precision corresponding to the singled-out
subject to test --i.e. compute the Gaussian log-likelihood of-- the data
of the left out session.

\begin{table}[b]
\begin{center}
{\small
\begin{tabular}{r|cccc|cccc|c}
& \multicolumn{4}{c|}{Using subject data} &
\multicolumn{4}{c|}{Uniform group model} &
\\
& MLE 	& LW & $\ell_2$  & $\ell_1$	 
& MLE 	& LW & $\ell_2$  & $\ell_1$  & $\ell_{21}$ 
\\
\hline
\\[-1em]
Generalization likelihood &
-57.1 & 33.1	& 38.8	    & 43.0	& 
40.6 & 41.5	& 41.6	    & 41.8 & 45.6
\\
Filling factor & 100\% & 100\% & 100\% & 45\% & 100\% & 100\% & 100\% & 60\% & 8\%
\\
Number of communities & 6 & 5 & 5 & 9 & 9 & 8 & 7 & 9 & 16
\\
Modularity & .07 & .07 & .12 & .25 & .23 & .23 & .18 & .32 & .60
\end{tabular}
}
\end{center}
\vspace*{-1em}
\caption{Summary statistics for different estimation strategies. MLE is
the Maximum Likelihood Estimate, in other words, the sample precision
matrix. LW is the Ledoit-Wolf estimate.
\label{tab:results}
}
\end{table}

\begin{figure}[p]
    \includegraphics[width=\linewidth]{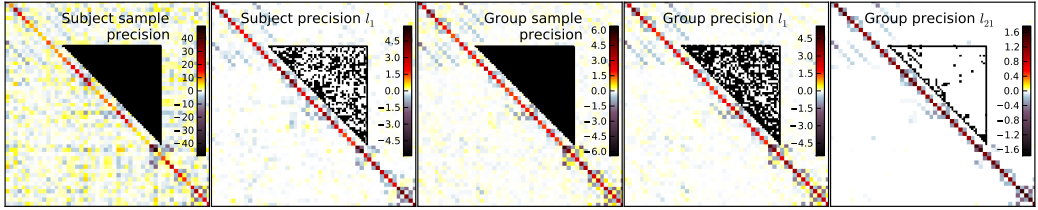}
    \vspace*{-2em}
    \caption{Precision matrices computed with different estimators. The
precision matrix is shown in false colors in the background and its
support is shown in black and white in an inset.
\label{fig:precisions}
}
\end{figure}

The cross-validation results (averaged across 20 subjects) are
reported in Table \ref{tab:results}. In addition, an example of
estimated precision matrices can be seen in Figure
\ref{fig:precisions}. We find that, due to the insufficient number of
samples in one session, the subject's sample precision matrix performs
poorly. $\ell_2$ penalization gives a good conditioning and better
performances, but is outperformed by $\ell_1$ penalized estimator that
yields a sparsity structure expressing conditional independences
between regions.
On the other hand, the population's sample precision is
well-conditioned due to the high number of samples at the group level
and generalizes much better than the subject-level sample precision or
the corresponding $\ell_2$-penalized estimate. Penalizing the
population-level covariance matrix does not give a significant
performance gain. In particular, the $\ell_1$-penalized subject-level
precision matrix outperforms the precision matrices learned from the
group ($p < 10^{-5}$).

We conclude from these cross-validation results that the generalization
power of the models estimated from the population data are not limited by
the number of samples but because they do not reflect the subject's
singularities. On the other hand, the estimation of a model solely from
the subject's data is limited by estimation error. We find that the
$\ell_{21}$-penalized estimator strikes a compromise and generalizes
significantly better than the other approaches ($p < 10^{-10}$). Although
each individual dataset is different and generalization scores vary from
subject to subject, compared to the second-best performing estimator the
$\ell_{21}$-penalized estimator gives a net gain for each subject of at
least 1.7 in the likelihood of unseen data.

\paragraph{Graphs estimated}
As can be seen from Figure \ref{fig:precisions}, precision matrices
corresponding to models that do not generalize well display a lot of
background noise whereas in models that generalize well, a sparse
structure stands out.
Although an $\ell_1$ penalization is sparsity inducing, the optimal
graphs estimated with such estimators are not very sparse (see table
\ref{tab:results}): a filling factor of 50\% amounts to $5\,000$ edges.
As a result, the corresponding graphs are not interpretable without
thresholding (corresponding visualization are given in the supplementary
materials). To interpret dense brain connectivity graphs, previous work
relied on extracting a connectivity backbone using a maximal spanning
tree \cite{hagmann2008}, or graph statistics on thresholded adjacency
matrices \cite{bullmore2009}.

On the opposite, the $\ell_{21}$-penalized graph is very sparse, with
only 700 edges. Adequate penalization serves as a replacement to
backbone extraction; moreover it corresponds to a theoretically well-grounded
and accurate model of brain connectivity.
After embedding in 3D anatomical space, the estimated graph is very
symmetric (see Figure \ref{fig:l21_graph}). A third of the weight on
the edges is on connections between a region and the corresponding one
on the opposite hemisphere.
In addition, the connectivity model displays strong fronto-parietal
connections, while the visual system is globally singled out into one
cluster, connected to the rest of the cortex mostly via the
middle-temporal area.


\section{An application: graph communities to describe functional
networks}

Even very sparse, high-dimensional functional connectivity graphs are
hard to interpret. However, they are deemed of high neuroscientific
interest, as their structure can reflect fundamental nervous system
assembly principles. 
Indeed, there is evidence from the study of the fault-resilient structure
of anatomical connections in the
nervous systems that ensembles of neurones cluster together to form
communities that are specialized to a cognitive task 
\cite{sporns2004,achard2006,hagmann2008}. 
This process, known as functional integration goes
along with a reduction of between-community connections, called
segregation.
So far, studies of full-brain connectivity graphs have focused on the
analysis of their statistical properties,
namely their small-world characteristics related to the emergence
of strongly-connected communities in neural system. These properties can be
summarized by a measure called {\sl modularity} 
\cite{achard2006,bullmore2009,newman2004}.
As the original measures introduced for
integration and segregation are Gaussian entropy and mutual
information measures \cite{tononi1994,sporns2000}, the estimation of a
well-conditioned Gaussian graphical model of the functional
signal gives us an adequate tool to study large-scale modularity and
integration in the brain. 
A limitation of the studies of statistical properties on graphs estimated
from the data is that they may reflect properties of the estimation noise.
Given that our graphical description generalizes well to unseen data, it
should reflect the intrinsic properties of brain functional
connectivity better than the sample correlation matrices previously used
\cite{achard2006}.
In this section, we study these properties on 
the optimal precision matrices describing a representative individual 
as estimated above.

\paragraph{Finding communities to maximize modularity}
Graph communities are a concept originally introduced in social
networks: communities are groups of densely-connected nodes with
little between-group connections. Newman and Girvan \cite{newman2004}
have introduced an objective function $Q$, called \textit{modularity},
to measure the quality of a graph partition in a community structure.
Choosing the partition to optimize modularity is a NP-hard problem,
but Smyth and White formulate it as a graph partitioning problem, and
give an algorithm \cite{smyth2005} based on a convex approximation
leading to spectral embedding and k-means clustering. The number of
classes is chosen to optimize modularity.

\begin{figure}[p]
\begin{minipage}{\linewidth}
    \hspace*{-.03\linewidth}%
    \includegraphics[height=.63\linewidth]{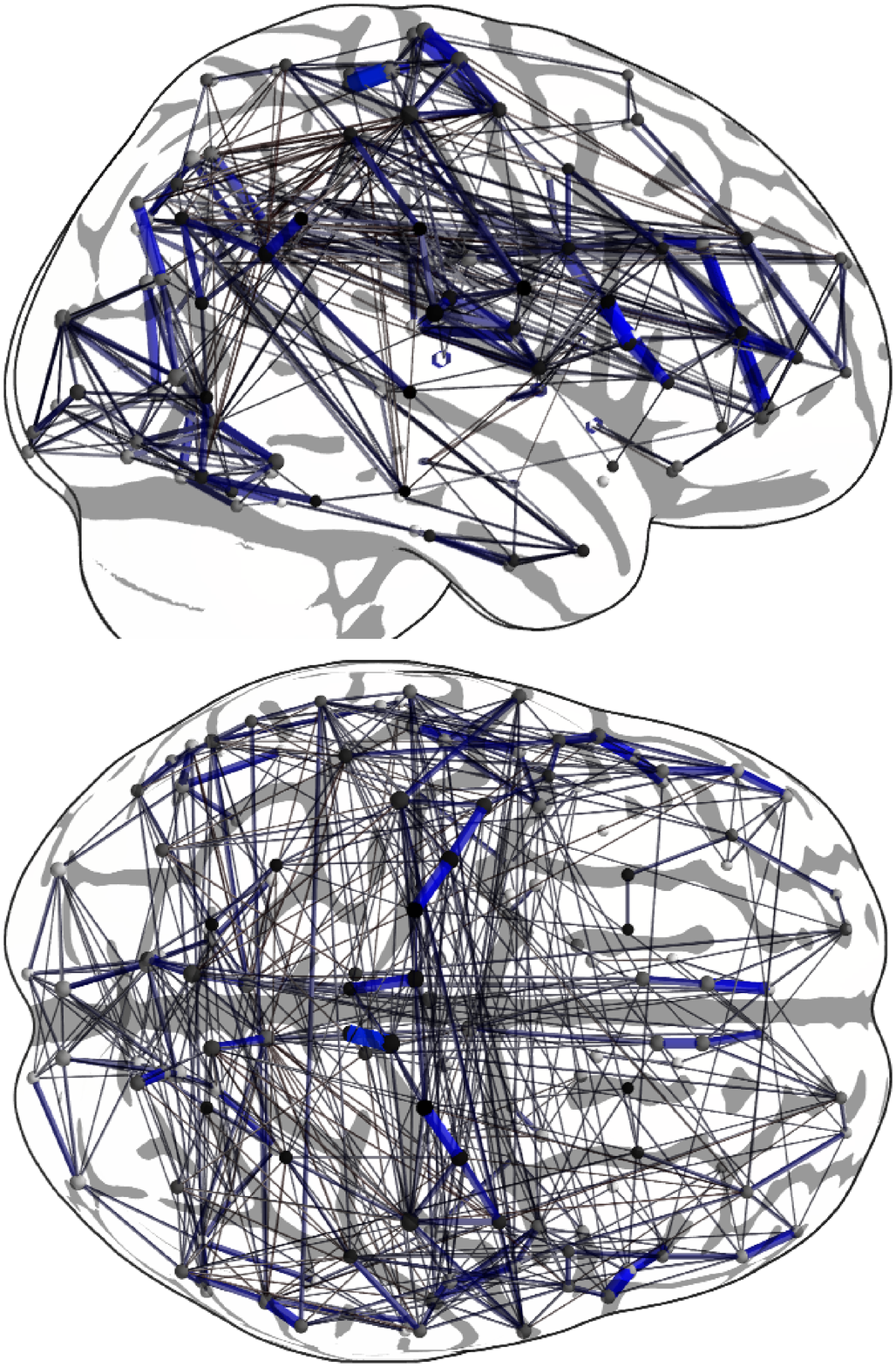}%
    \llap{\sffamily\bfseries 
	\raisebox{.35\linewidth}{\colorbox{white}{Full graph}} 
	\hspace*{15ex}}%
    \hfill%
    \includegraphics[height=.63\linewidth]{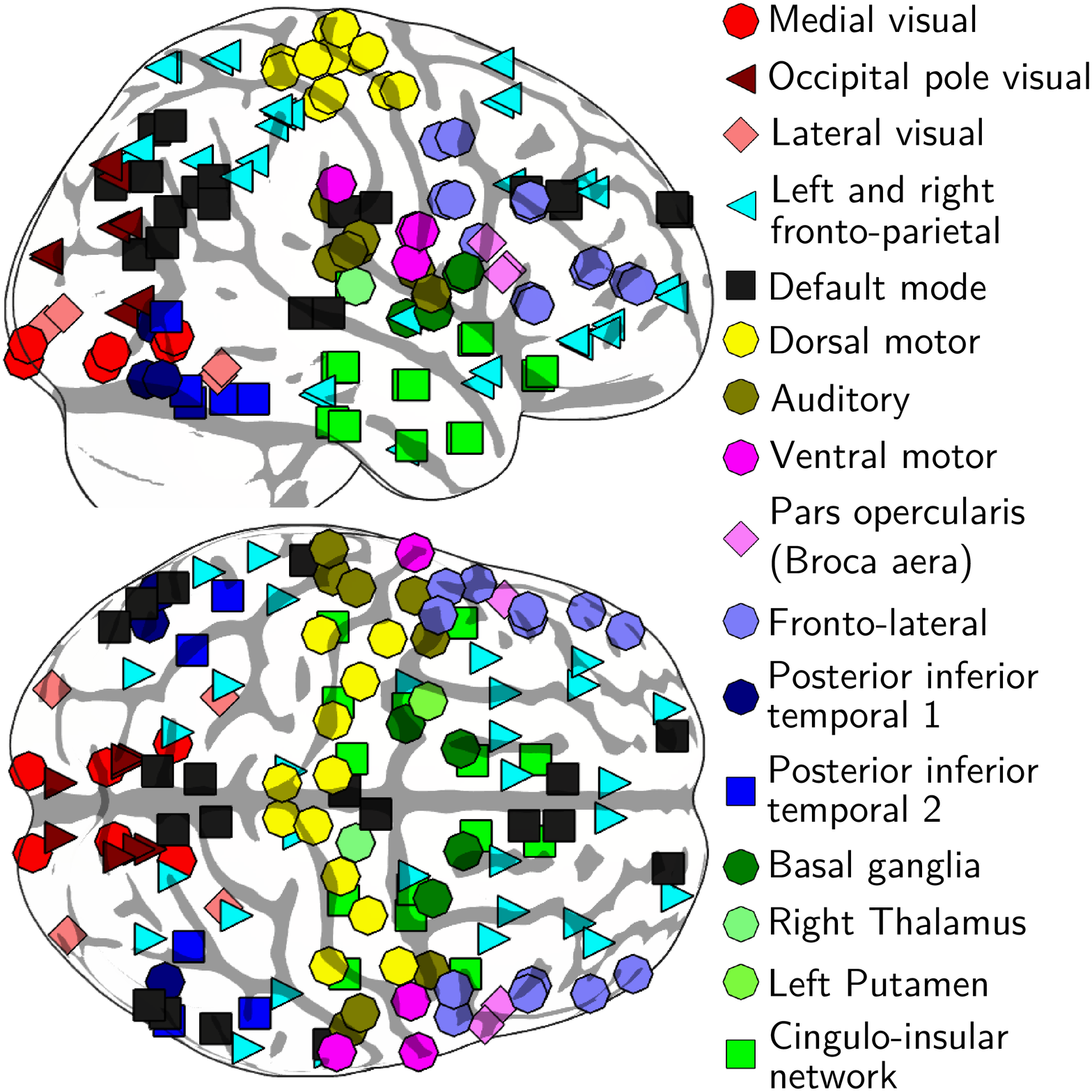}%
    \llap{\sffamily\bfseries 
	\raisebox{.34\linewidth}{\colorbox{white}{Communities}} 
	\hspace*{32ex}}%
\end{minipage}%
\vspace*{-.5em}
    \caption{Functional-connectivity graph computed by
$\ell_{21}$-penalized estimation and corresponding communities. The graph
displayed on the left is not thresholded, but on the top view, 
connections linking
one region to its corresponding one on the opposite hemisphere are not
displayed.
\label{fig:l21_graph}
}
\end{figure}

\paragraph{Brain functional-connectivity communities}
We apply Smyth and White's algorithm on the brain connectivity graphs. We
find that using the $\ell_{21}$-penalized precision matrices yields a
higher number of communities, and higher modularity values (Table
\ref{tab:results}) then the other estimation strategies. We discuss in
details the results obtained without regularization, and with the best
performing regularization strategies: $\ell_1$ penalization on individual
data, and $\ell_{21}$ penalization. 
The communities extracted from the sample precision matrix are mostly
spread throughout the brain, while the graph estimated with $\ell_1$
penalization on individual data yields communities centered on
anatomo-functional regions such as the visual system 
(figures in supplementary materials).
The communities extracted on the $\ell_{21}$-penalized precision
exhibit finer anatomo-functional structures, but also extract some
known functional networks that are commonly found while studying
spontaneous as well as task-related activity \cite{smith2009}.
In Figure \ref{fig:l21_graph}, we display the resulting communities,
making use, when possible, of the same denominations as the functional
networks described in \cite{smith2009}.
In particular, the {\sl default mode} network and the fronto-parietal
network are
structures reproducibly found in functional-connectivity studies that
are non-trivial as they are large-scale, long-distance, and not
comprised solely of bilateral regions.

\begin{figure}[p]
    \vspace*{-1ex}%
    \includegraphics[height=.3\linewidth]{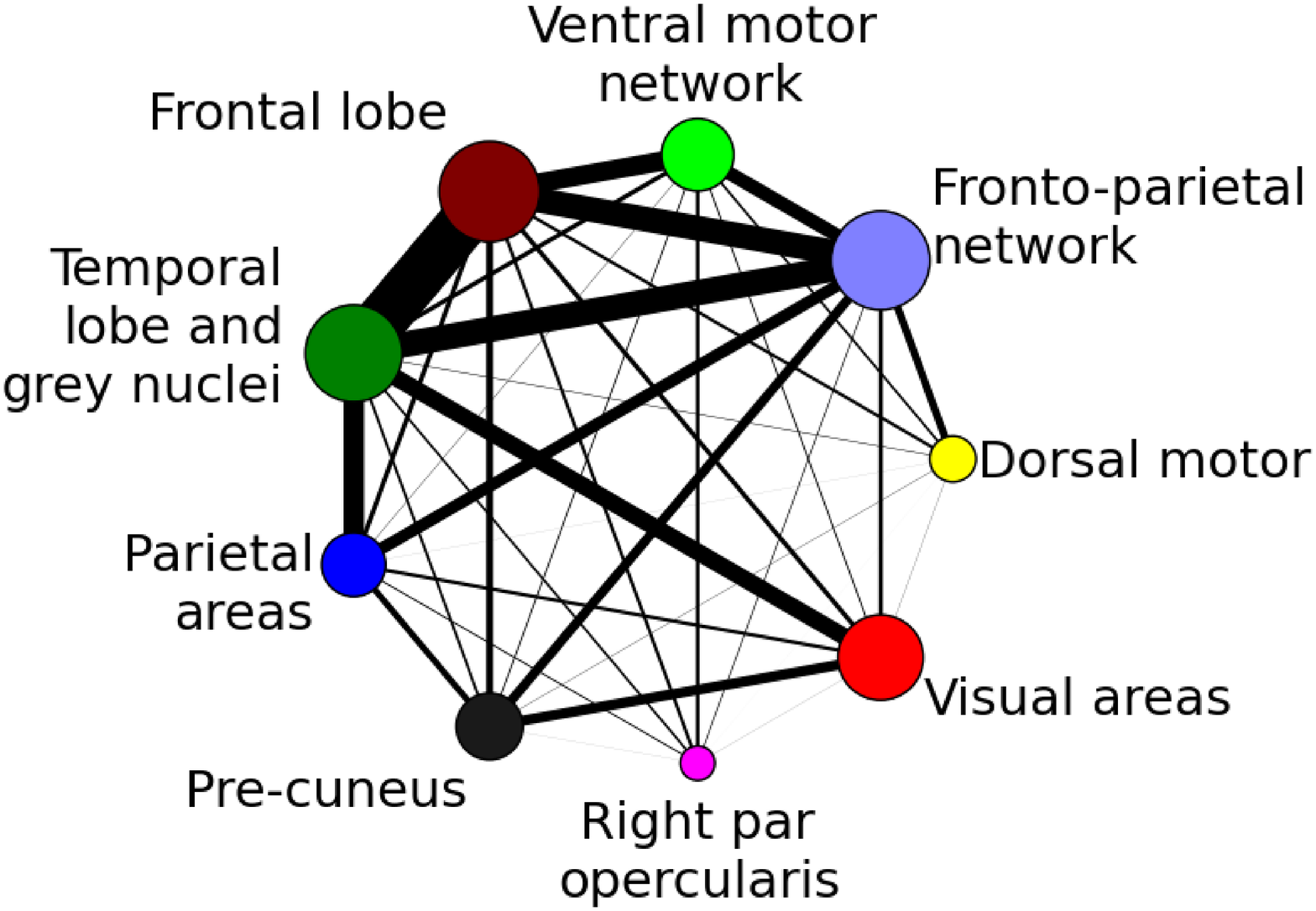}%
    \llap{\sffamily\bfseries 
  \raisebox{1ex}{\large $\ell_{1}$}
  \hspace*{1ex}}%
    \hfill%
    \includegraphics[height=.31\linewidth]{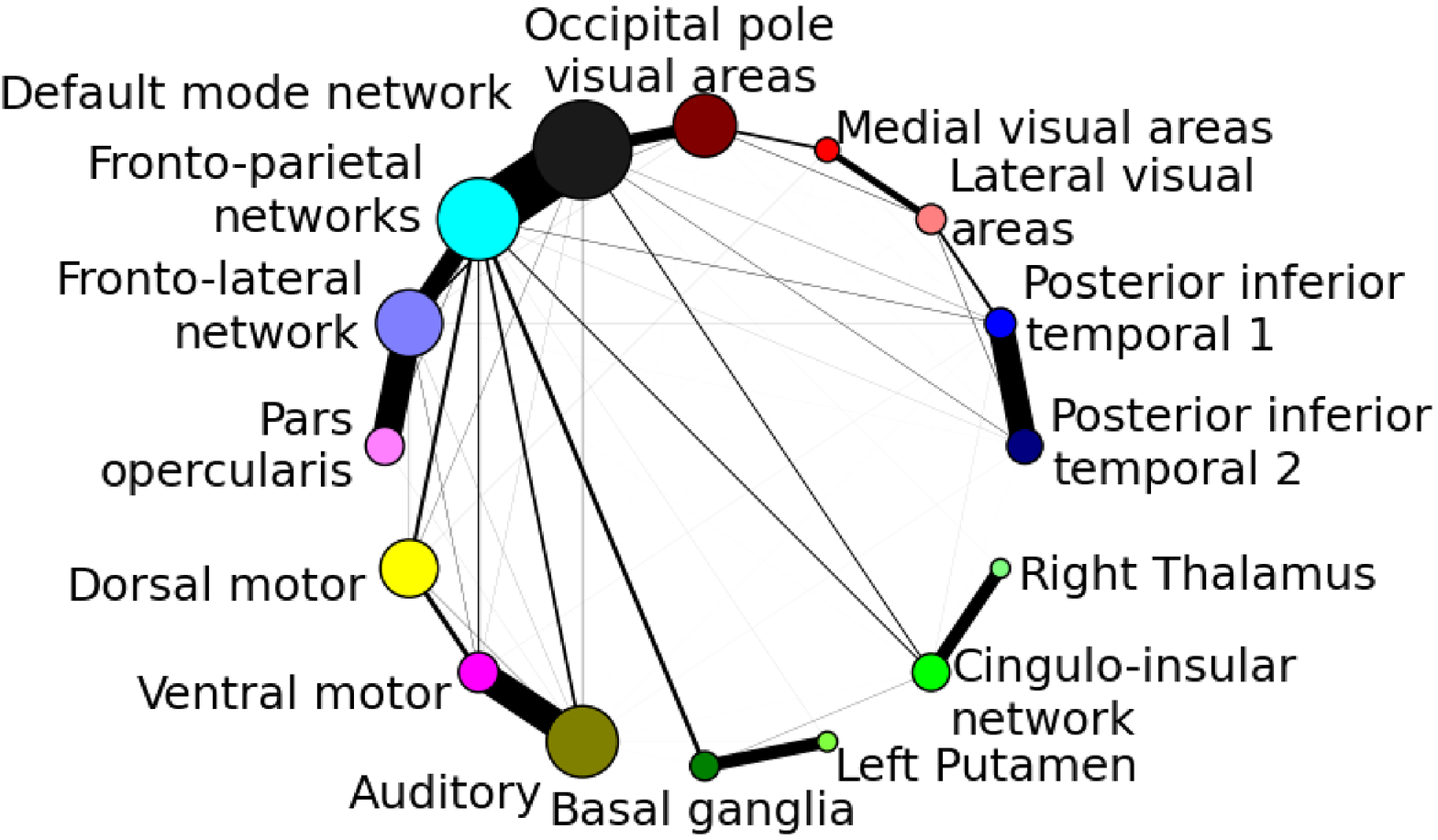}%
    \llap{\sffamily\bfseries 
  \raisebox{1ex}{\large $\ell_{21}$}
  \hspace*{1ex}}%
\vspace*{-1em}
    \caption{Between-communities integration graph obtained through
    $\ell_{1}$- (left) and $\ell_{21}$-penalization (right). The size
    of the nodes represents integration within a community and the
    size of the edges represents mutual information between
    communities. Region order is chosen via 1D Laplace embedding. The
    regions comprising the communities for the $\ell_{1}$-penalized
    graph are detailed in the supplementary materials.
\label{fig:integration_graph}
}
\end{figure}

\paragraph{Integration and segregation in the graph communities}
These functionally-specialized networks are thought to be the expression of integration and
segregation processes in the brain circuits architecture. We apply the 
measures introduced by Tononi {\sl et al.} \cite{tononi1994} on the 
estimated
graphs to quantify this integration and segregation, namely Gaussian
entropy of the functional networks, and mutual information. 
However, following \cite{coynel2010}, we use conditional integration
and conditional mutual information to obtain conditional pair-wise
measures, and thus a sparser graph: for two sets of nodes $S_1$ and
$S_2$, 
\vspace*{-1ex}%
\begin{eqnarray}
\text{Integration:} & I_{S_1} & = \frac{1}{2}\log\det ( \B{K}_{S_1} )
\\
\text{Mutual information:} & M_{S_1, S_2} & = I_{S_1 \cup S_2} - I_{S_1} -
I_{S_2},
\end{eqnarray}
%
%
where $\B{K}_{S_1}$ denotes the precision matrix restricted to the nodes
in $S_1$. We use these two measures, pair-wise and within-community, to
create a graph between communities.

This graph reflects the large-scale brain function organization. We
compare the graph built using the $\ell_1$ and $\ell_{21}$-penalized
precisions (figure \ref{fig:integration_graph}). 
We find that the former is much sparser than the latter,
reflecting a higher large segregation in between the communities
estimated. The graph corresponding to the $\ell_{21}$ penalization
segments the brain in smaller communities and care must be taken in
comparing the relative integration of the different systems: for instance
the visual system appears as more integrate on the $\ell_{1}$ graph, but
this is because it is split in three on the $\ell_{21}$ graph.

Although this graph is a very simplified view of brain
functional architecture at rest, it displays some of the key
processing streams: starting from the primary visual system (medial
visual areas), we can distinguish the dorsal visual pathway, going
through the occipital pole to the intra-parietal areas comprised in
the default mode network and the fronto-parietal networks, as well as
the ventral visual pathway, going through the lateral visual areas to
the inferior temporal lobe. 
The default mode  and the fronto-parietal networks appear as hubs,
connecting different networks with different functions, such as the
visual streams, but also the motor areas, as well as the frontal
regions.



\section{Conclusion}

We have presented a strategy to overcome the challenge of
subject-to-subject variability and learn a detailed model of an
individual's full-brain functional connectivity using population data.
The learnt graphical model is sparse and reveals the interaction
structure between functional modules via conditional independence
relationships that generalize to new data. 
As far as we can tell, this is the first time an unsupervised model of
brain functional connectivity is backed by cross-validation. Also,
from a machine learning perspective, this work is the first
demonstration, to our knowledge, of joint estimation of multiple graphical 
models
in a model-selection setting, and the first time it is shown to
improve a prediction score for individual graphical models.

From a neuroscience perspective, learning high-dimensional
functional connectivity probabilistic models opens the door to new
studies of brain architecture. 
In particular, the models estimated with our strategy are well suited
to exploring the graph-community structure resulting from the
functional integration, specialization, and segregation of distributed
networks. Our preliminary work suggests that a mesoscopic description
of neural ensembles via high-dimensional graphical models can
establish the link between the functional networks observed in brain
imaging and the fundamental nervous-system assembly principles.
%
Finally, subject-level Gaussian probabilistic models of functional
connectivity between a few regions have proved useful for
statistically-controlled inter-individual comparisons on resting-state,
with medical applications \cite{varoquaux2010b}. Extending such studies to
full-brain analysis, that have been so-far 
limited by the amount of data
available on individual subjects, clears the way to new insights in brain pathologies
\cite{cecchi2009,huang2009}.


{
\small
\bibliographystyle{my_splncs}
\bibliography{restingstate}

\begin{thebibliography}{10}

\bibitem{Fox2007}
M.~Fox and M.~Raichle:
\newblock Spontaneous fluctuations in brain activity observed with functional
  magnetic resonance imaging.
\newblock Nat Rev Neurosci \textbf{8} (2007)  700--711

\bibitem{bullmore2009}
E.~Bullmore and O.~Sporns:
\newblock Complex brain networks: graph theoretical analysis of structural and
  functional systems.
\newblock Nat Rev Neurosci \textbf{10} (2009)  186--198

\bibitem{smith2009}
S.~Smith, et~al. :
\newblock {Correspondence of the brain's functional architecture during
  activation and rest}.
\newblock PNAS \textbf{106} (2009)  13040

\bibitem{achard2006}
S.~Achard, et~al. :
\newblock {A resilient, low-frequency, small-world human brain functional
  network with highly connected association cortical hubs}.
\newblock J Neurosci \textbf{26} (2006) ~63

\bibitem{sporns2004}
O.~Sporns, et~al. :
\newblock {Organization, development and function of complex brain networks}.
\newblock Trends in Cognitive Sciences \textbf{8} (2004)  418--425

\bibitem{cecchi2009}
G.~Cecchi, et~al. :
\newblock Discriminative network models of schizophrenia.
\newblock In: NIPS 22.
\newblock (2009)  250--262

\bibitem{seeley2009}
W.~Seeley, et~al. :
\newblock {Neurodegenerative Diseases Target Large-Scale Human Brain Networks}.
\newblock Neuron \textbf{62} (2009)  42--52

\bibitem{huang2009}
S.~Huang, et~al. :
\newblock Learning brain connectivity of {Alzheimer's} disease from
  neuroimaging data.
\newblock In: Advances in Neural Information Processing Systems 22.
\newblock (2009)  808--816

\bibitem{varoquaux2010b}
G.~Varoquaux, et~al. :
\newblock Detection of brain functional-connectivity difference in post-stroke
  patients using group-level covariance modeling.
\newblock In: IEEE MICCAI.
\newblock (2010)

\bibitem{greicius2008b}
M.~Greicius:
\newblock {Resting-state functional connectivity in neuropsychiatric
  disorders}.
\newblock Current opinion in neurology \textbf{21} (2008)  424

\bibitem{McIntosh1994}
A.~McLntosh and F.~Gonzalez-Lima:
\newblock Structural equation modeling and its application to network analysis
  in functional brain imaging.
\newblock Human Brain Mapping \textbf{2}(1) (1994)  2--22

\bibitem{daunizeau2009}
J.~Daunizeau, K.~Friston, and S.~Kiebel:
\newblock {Variational Bayesian identification and prediction of stochastic
  nonlinear dynamic causal models}.
\newblock Physica D \textbf{238} (2009)

\bibitem{Honorio2010}
J.~Honorio and D.~Samaras:
\newblock {Multi-Task Learning of Gaussian Graphical Models}.
\newblock In: ICML. (2010)

\bibitem{Dempster1972}
A.~Dempster:
\newblock Covariance selection.
\newblock Biometrics \textbf{28}(1) (1972)  157--175

\bibitem{banerjee2006}
O.~Banerjee, et~al. :
\newblock {Convex optimization techniques for fitting sparse Gaussian graphical
  models}.
\newblock In: ICML. (2006) ~96

\bibitem{duchi2008}
J.~Duchi, S.~Gould, and D.~Koller:
\newblock {Projected subgradient methods for learning sparse gaussians}.
\newblock In: Proc. of the Conf. on Uncertainty in AI. (2008)

\bibitem{Friedman2008}
J.~Friedman, T.~Hastie, and R.~Tibshirani:
\newblock Sparse inverse covariance estimation with the graphical lasso.
\newblock Biostatistics \textbf{9}(3) (2008)  432--441

\bibitem{yuan2006}
M.~Yuan and Y.~Lin:
\newblock {Model selection and estimation in regression with grouped
  variables}.
\newblock Journal-Royal Statistical Society Series B Statistical Methodology
  \textbf{68}(1) (2006) ~49

\bibitem{guo2009}
J.~Guo, et~al. :
\newblock Joint estimation of multiple graphical models.
\newblock Preprint (2009)

\bibitem{chiquet2010}
J.~Chiquet, Y.~Grandvalet, and C.~Ambroise:
\newblock Inferring multiple graphical structures.
\newblock Stat and Comput (2010)

\bibitem{Rothman2008}
A.~Rothman, et~al. :
\newblock Sparse permutation invariant covariance estimation.
\newblock Electron J Stat \textbf{2} (2008)  494

\bibitem{ledoit2004}
O.~Ledoit and M.~Wolf:
\newblock A well-conditioned estimator for large-dimensional covariance
  matrices.
\newblock J. Multivar. Anal. \textbf{88} (2004)  365--411

\bibitem{beckmann2004}
C.~F. Beckmann and S.~M. Smith:
\newblock Probabilistic independent component analysis for functional magnetic
  resonance imaging.
\newblock Trans Med Im \textbf{23}(2) (2004)  137--152

\bibitem{perrot2009}
M.~Perrot, et~al. :
\newblock {Joint Bayesian Cortical Sulci Recognition and Spatial
  Normalization}.
\newblock In: {{IPMI}}. (2009)

\bibitem{keller2009}
M.~Keller, et~al. :
\newblock {Anatomically Informed Bayesian Model Selection for fMRI Group Data
  Analysis}.
\newblock In: {{MICCAI}}. (2009)

\bibitem{cordes2000}
D.~Cordes, et~al. :
\newblock {Mapping functionally related regions of brain with functional
  connectivity MR imaging}.
\newblock American Journal of Neuroradiology \textbf{21}(9) (2000)  1636--1644

\bibitem{hagmann2008}
P.~Hagmann, et~al. :
\newblock Mapping the structural core of human cerebral cortex.
\newblock PLoS Biol \textbf{6}(7) (2008)  e159

\bibitem{newman2004}
M.~Newman and M.~Girvan:
\newblock {Finding and evaluating community structure in networks}.
\newblock Phys rev E \textbf{69} (2004)  26113

\bibitem{tononi1994}
G.~Tononi, O.~Sporns, and G.~Edelman:
\newblock {A measure for brain complexity: relating functional segregation and
  integration in the nervous system}.
\newblock PNAS \textbf{91} (1994)  5033

\bibitem{sporns2000}
O.~Sporns, G.~Tononi, and G.~Edelman:
\newblock {Theoretical neuroanatomy: relating anatomical and functional
  connectivity in graphs and cortical connection matrices}.
\newblock Cereb Cortex \textbf{10} (2000)  127

\bibitem{smyth2005}
S.~White and P.~Smyth:
\newblock {A spectral clustering approach to finding communities in graphs}.
\newblock In: 5th SIAM international conference on data mining. (2005)  274

\bibitem{coynel2010}
D.~Coynel, et~al. :
\newblock Conditional integration as a way of measuring mediated interactions
  between large-scale brain networks in functional {MRI}.
\newblock In: Proc. ISBI. (2010)

\end{thebibliography}
}

\clearpage

\section*{Supplementary materials}

\begin{figure}[h]
    \includegraphics[height=.6\linewidth]{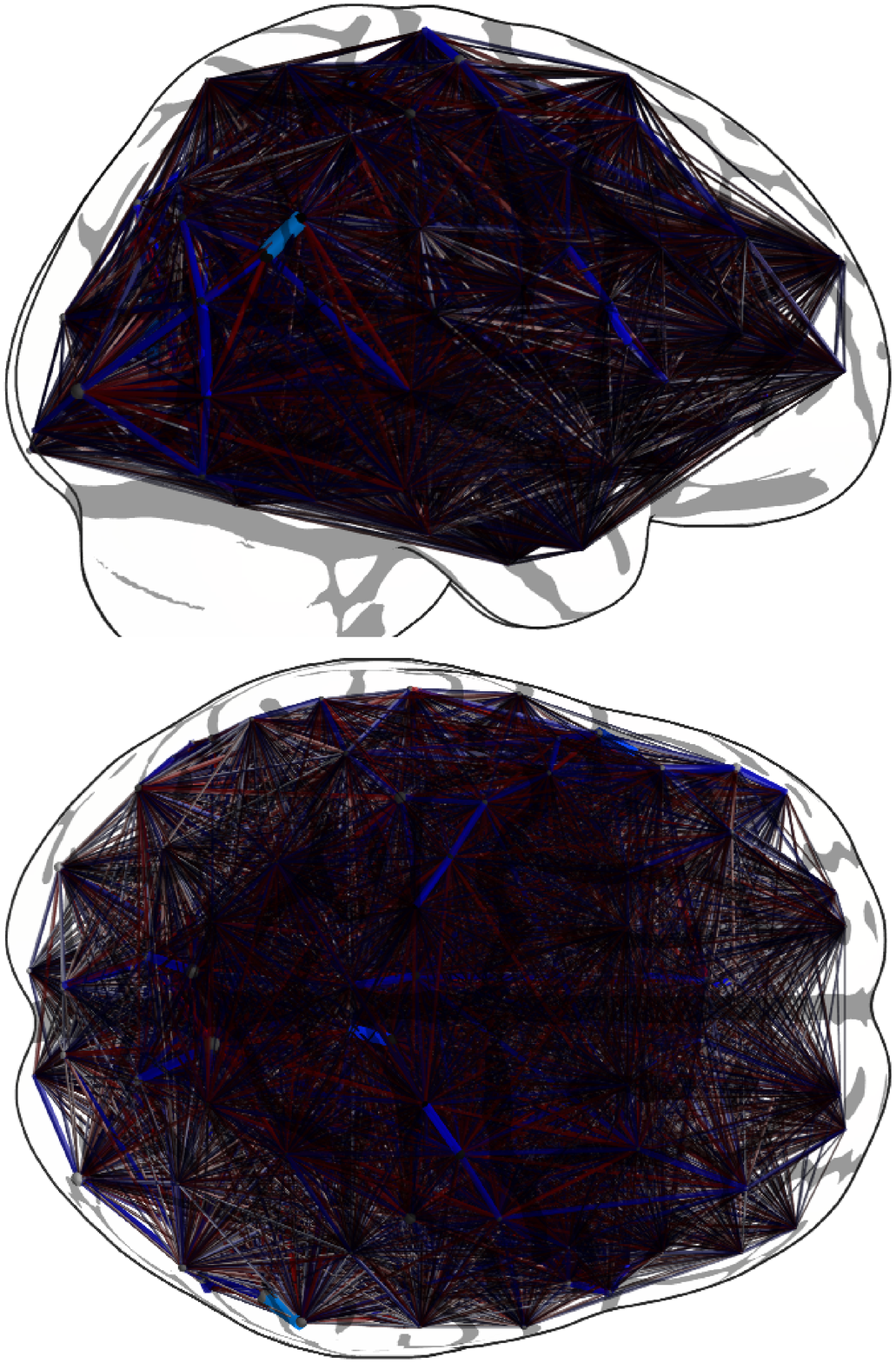}%
    \llap{\sffamily\bfseries 
	\raisebox{.33\linewidth}{\colorbox{white}{Full graph}} 
	\hspace*{15ex}}%
    \hfill%
    \includegraphics[height=.6\linewidth]{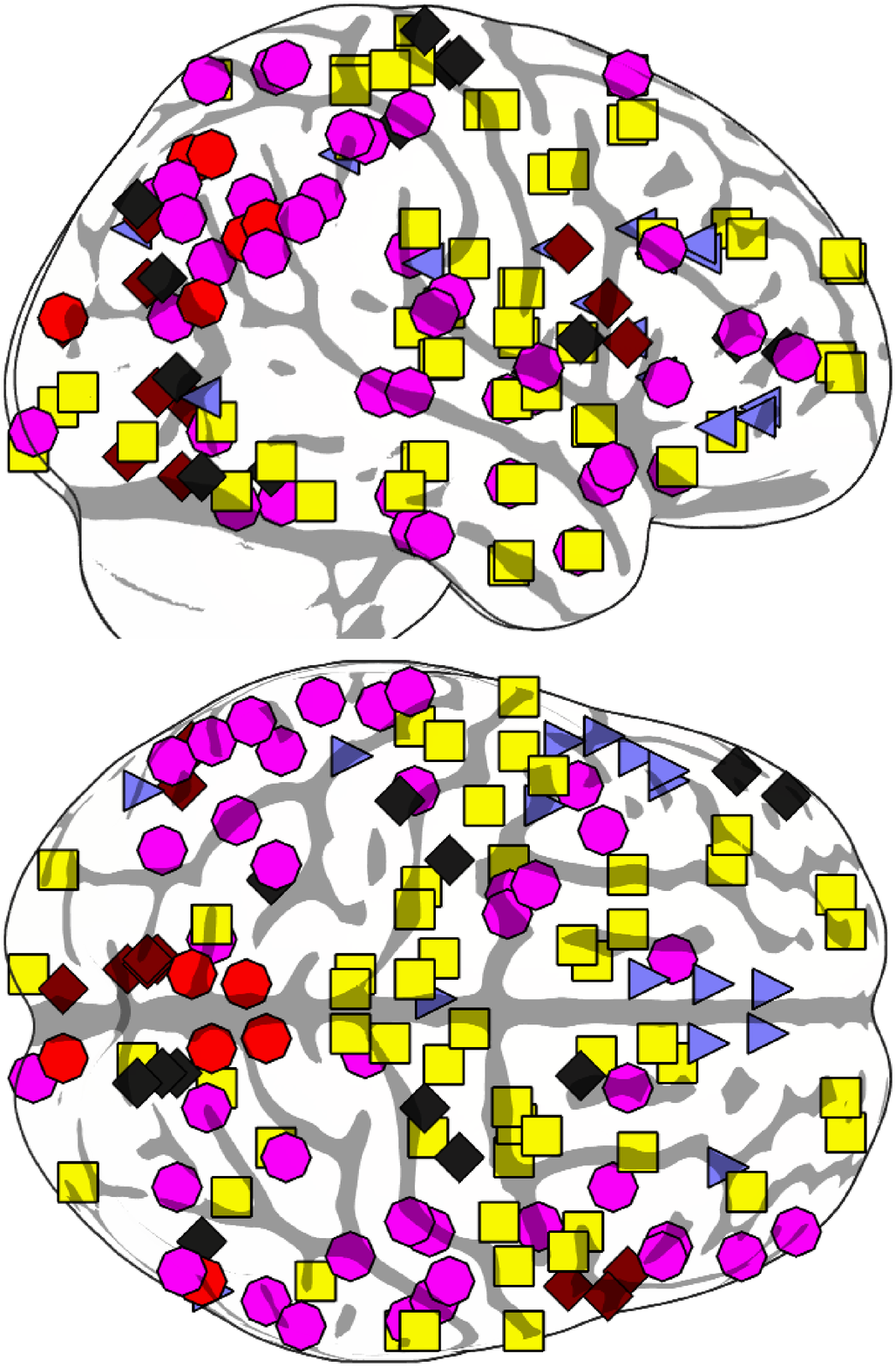}
    \llap{\sffamily\bfseries 
	\raisebox{.14\linewidth}{%
	\includegraphics[width=.2\linewidth]{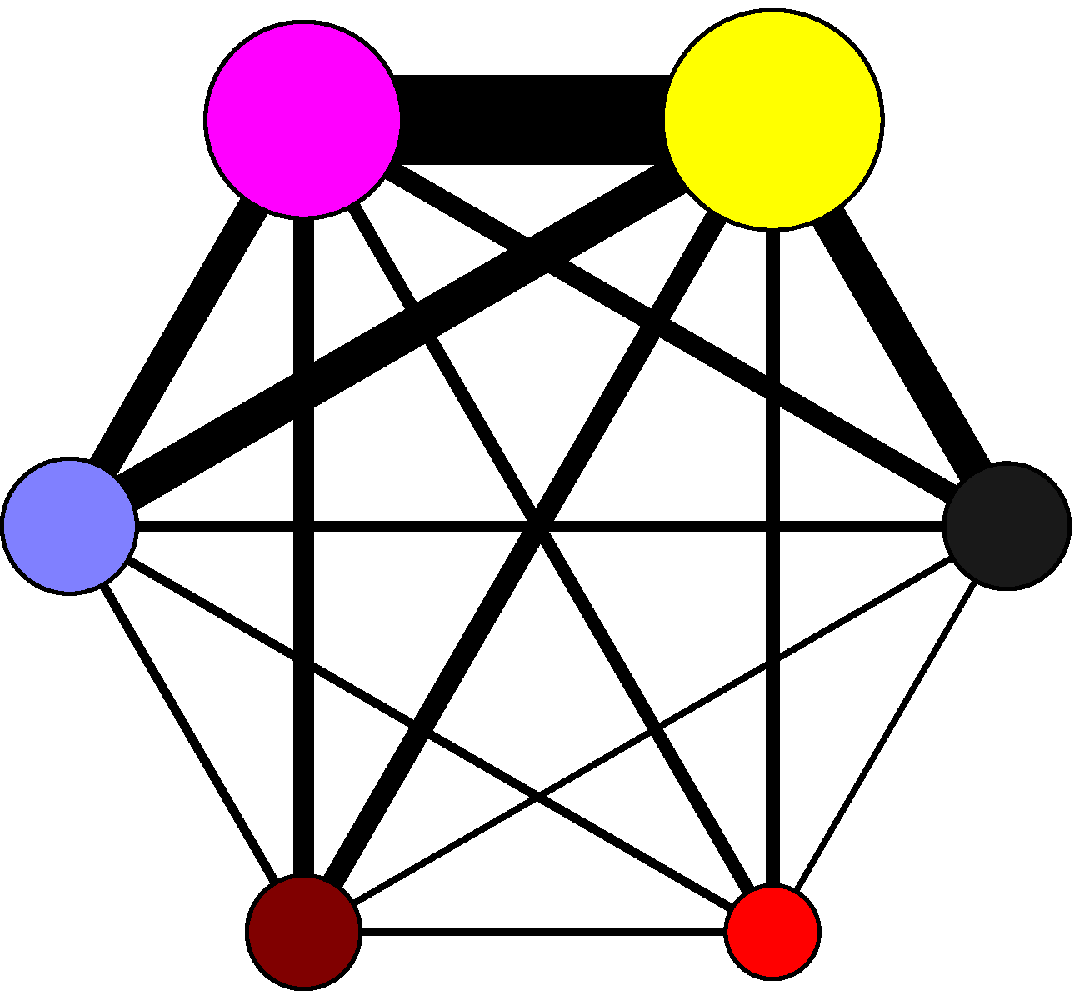}%
	}
	\hspace*{30ex}}%
    \llap{\sffamily\bfseries 
	\raisebox{.36\linewidth}{%
	\sffamily\bfseries Between-communities
	}
	\hspace*{28ex}}%
    \llap{\sffamily\bfseries 
	\raisebox{.33\linewidth}{%
	\sffamily\bfseries integration graph
	}
	\hspace*{30ex}}%
    \llap{\sffamily\bfseries 
	\raisebox{.32\linewidth}{\colorbox{white}{Communities}} 
	\hspace*{13ex}}%
    
    \caption{Graph computed by maximum likelihood estimation on an
individual subject's data. The graph displayed is not thresholded, but
on the top view, connections linking one region to its corresponding one
on the opposite hemisphere are not displayed.}
\end{figure}

\begin{figure}[h]
    \includegraphics[height=.6\linewidth]{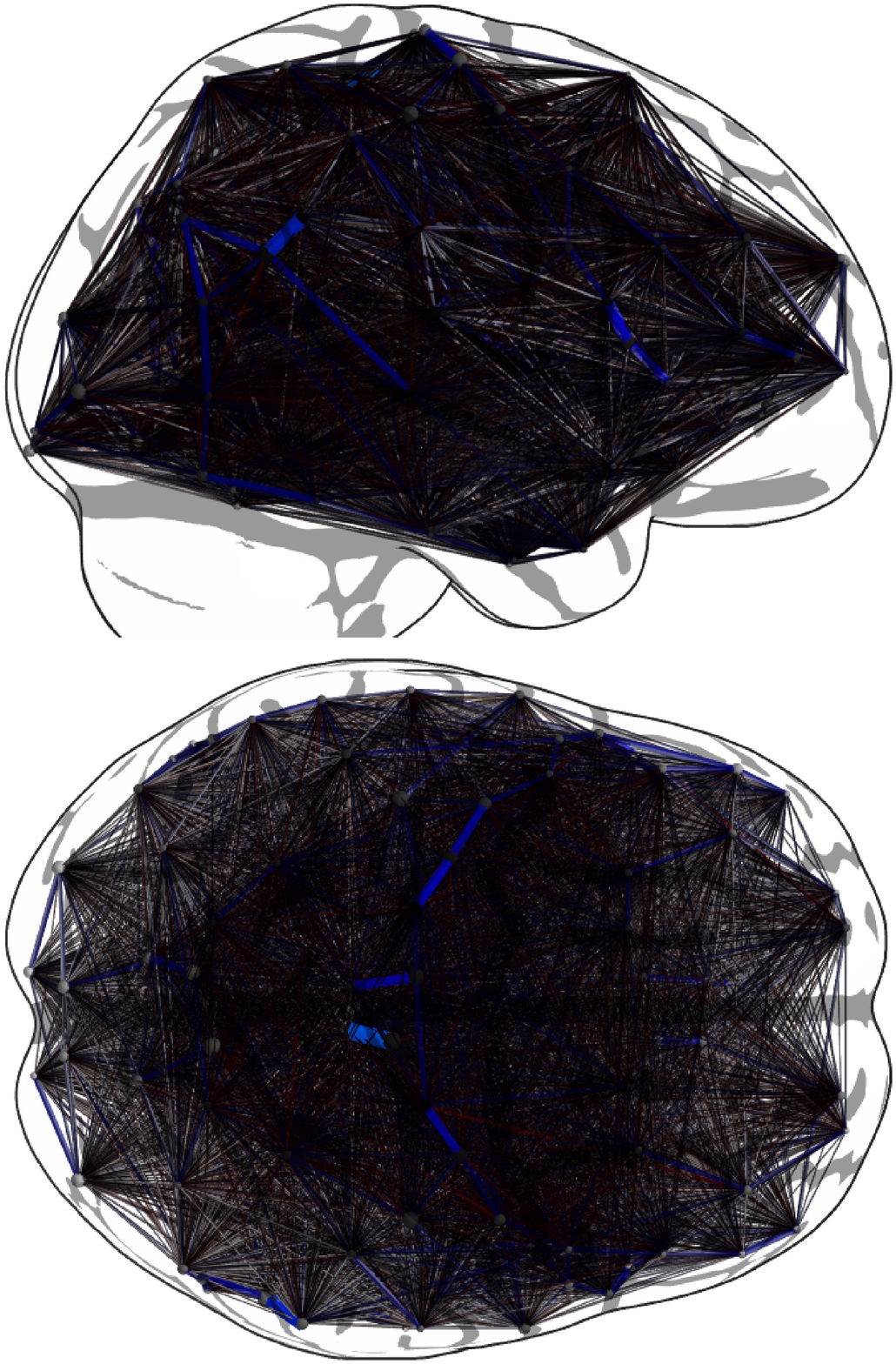}%
    \llap{\sffamily\bfseries 
	\raisebox{.33\linewidth}{\colorbox{white}{Full graph}} 
	\hspace*{15ex}}%
    \hfill%
    \includegraphics[height=.6\linewidth]{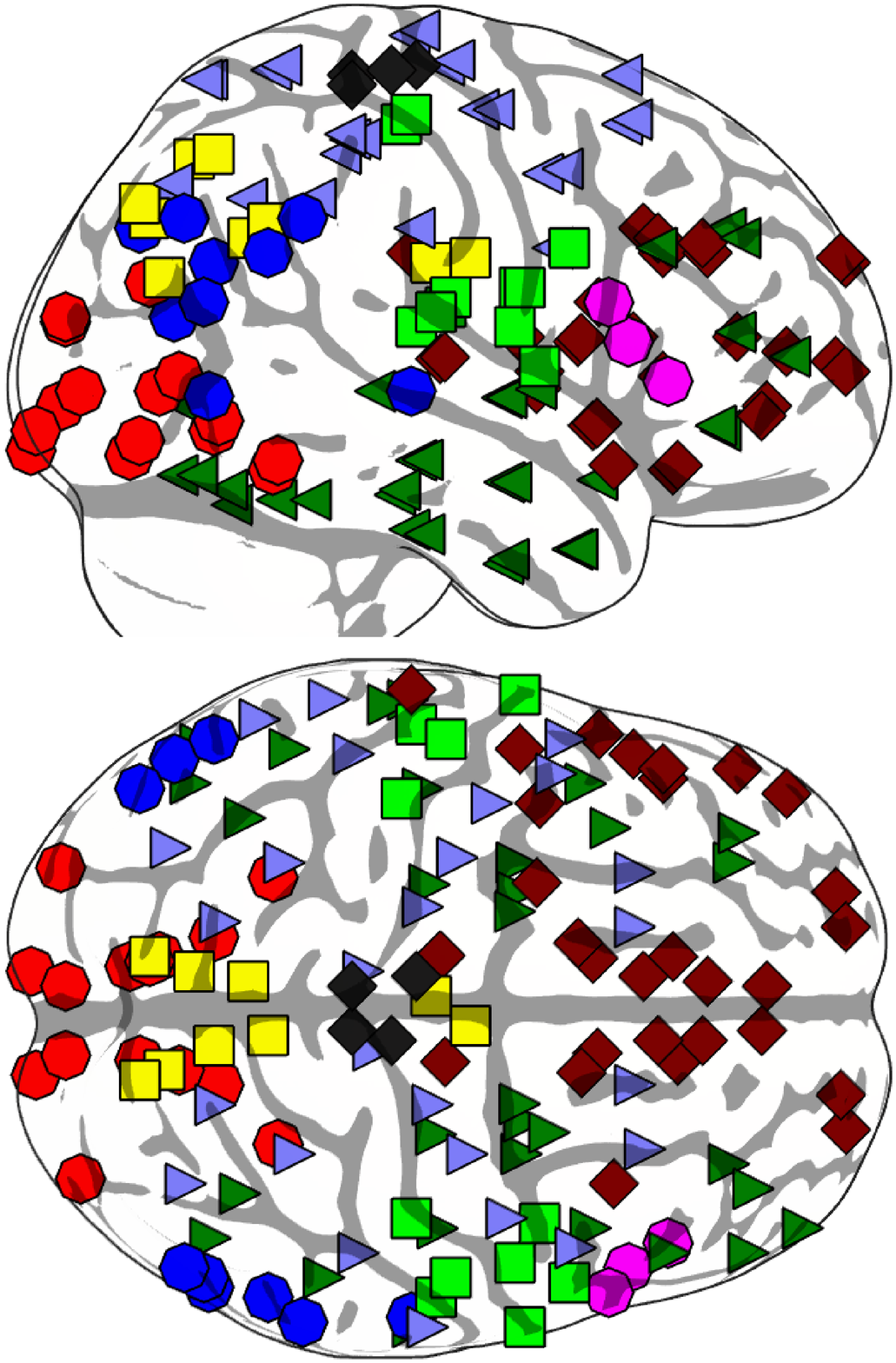}
    \llap{\sffamily\bfseries 
	\raisebox{.32\linewidth}{\colorbox{white}{Communities}} 
	\hspace*{13ex}}%
    
    \caption{Graph computed by $\ell_1$-penalized estimation on an
    individual subject's data. The graph displayed is not thresholded, but
on the top view, connections linking one region to its corresponding one
on the opposite hemisphere are not displayed.}
\end{figure}

\end{document}